\documentclass[a4paper]{article}

\usepackage{itgspeech2023}    %
\usepackage{times}            %
\usepackage[english]{babel}   %
\usepackage[ansinew]{inputenc}%
\usepackage[T1]{fontenc}      %
\usepackage[sort&compress,numbers]{natbib}	%
\usepackage{amsmath,amssymb}
\usepackage{graphicx}
\usepackage{microtype}
\usepackage{setspace}
\usepackage[colorlinks=false,pdfborder={0 0 0}]{hyperref}
\usepackage[capitalize]{cleveref}
\usepackage[acronym,toc,shortcuts,nonumberlist]{glossaries}
\usepackage{multirow}
\usepackage{siunitx}
\usepackage{xspace}

\newacronym{AM}{AM}{acoustic model}
\newacronym{AMI}{AMI}{Augmented Multi-party Interaction}
\newacronym{ARQ}{ARQ}{automatic repeat request}
\newacronym{ASR}{ASR}{automatic speech recognition}
\newacronym[longplural={bi-directional long-short term memories}]{BLSTM}{BLSTM}{bi-directional long-short term memory}
\newacronym{BSS}{BSS}{blind speech separation}
\newacronym{CART}{CART}{classification and regression tree}
\newacronym{CE}{CE}{cross entropy}
\newacronym{CER}{CER}{character error rate}
\newacronym{CDp}{CDp}{context dependent phoneme}
\newacronym{CNN}{CNN}{convolutional neural network}
\newacronym{CPC}{CPC}{contrastive predictive coding}
\newacronym{CTC}{CTC}{connectionist temporal classification}
\newacronym{DCT}{DCT}{discrete cosine transform}
\newacronym{DL}{DL}{deep learning}
\newacronym{DNN}{DNN}{deep neural network}
\newacronym{DNN-HMM}{DNN-HMM}{deep neural network hidden Markov model}
\newacronym{ELBO}{ELBO}{evidence lower bound}
\newacronym{EM}{EM}{expectation maximization}
\newacronym{FE}{FE}{feature extractor}
\newacronym{FIR}{FIR}{finite impulse response}
\newacronym{FFNN}{FFNN}{feed-forward neural network}
\newacronym{fCE}{fCE}{frame-wise cross-entropy}
\newacronym{G2P}{G2P}{grapheme-to-phoneme conversion}
\newacronym{GAN}{GAN}{generative adversarial network}
\newacronym{GMM}{GMM}{Gaussian mixture model}
\newacronym{GMM-HMM}{GMM-HMM}{Gaussian mixture model hidden Markov model}
\newacronym{GPU}{GPU}{graphics processing unit}
\newacronym{HMM}{HMM}{hidden Markov model}
\newacronym{IHM}{IHM}{individual headset microphones}
\newacronym{IIR}{IIR}{infinite impulse response}
\newacronym{LAS}{LAS}{listen-attend-spell}
\newacronym{LC-BLSTM}{LC-BLSTM}{latency-controlled bidirectional long-short term memory}
\newacronym{LDA}{LDA}{linear discriminant analysis}
\newacronym{LM}{LM}{language model}
\newacronym[longplural={long-short term memories}]{LSTM}{LSTM}{long-short term memory}
\newacronym{MDM}{MDM}{multiple distant microphones}
\newacronym{MFCC}{MFCC}{Mel-frequency cepstral coefficients}
\newacronym{MHSA}{MHSA}{multi-head self-attention}
\newacronym{MRES}{MRES}{multi-resolutional}
\newacronym{MSA}{MSA}{Modern Standard Arabic}
\newacronym{MT}{MT}{machine translation}
\newacronym{NLP}{NLP}{natural language processing}
\newacronym{NN}{NN}{neural network}
\newacronym{OOV}{OOV}{out-of-vocabulary}
\newacronym{PC2}{$\text{PC}^{\text{2}}$}{Paderborn Center for Parallel Computing}
\newacronym{PPL}{PPL}{perplexity}
\newacronym{RNA}{RNA}{recurrent neural aligner}
\newacronym{RNN}{RNN}{recurrent neural network}
\newacronym{RSAN}{RSAN}{recurrent selective attention network}
\newacronym{SAT}{SAT}{speaker adaptive training}
\newacronym{SDM}{SDM}{single distant microphone}
\newacronym{SDR}{SDR}{signal-to-distortion ratio}
\newacronym{SC}{SC}{supervised convolutional}
\newacronym{sMBR}{sMBR}{state-level minimum Bayes risk}
\newacronym{STFT}{STFT}{short time Fourier transform}
\newacronym{TDNN}{TDNN}{time delay neural network}
\newacronym[longplural={time-frequencies}]{tf}{tf}{time-frequency}
\newacronym{VAD}{VAD}{voice activity detection}
\newacronym{VTLN}{VTLN}{vocal tract length normalization}
\newacronym{cpWER}{cpWER}{concatenated minimum-permutation word error rate}
\newacronym{WER}{WER}{word error rate}
\newacronym{WERR}{WERR}{word error rate reduction}
\newacronym{WP}{WP}{work package}
\newacronym{WPE}{WPE}{weighted prediction error}
\newacronym{WSJ}{WSJ}{Wall Street Journal}

\newcommand{\tab}{Table~}

\newcommand{\sect}{Section~}

\newcommand{\transformer}{\textit{Transformer}\xspace}
\newcommand{\conformer}{\textit{Conformer}\xspace}

\title{Development of Hybrid ASR Systems for Low Resource\\Medical Domain Conversational Telephone Speech}

\author{Christoph L\"uscher$^{1,2}$, Mohammad Zeineldeen$^{1,2}$, Zijian Yang$^{1,2}$, Tina Raissi$^{1}$, Peter Vieting$^{1}$,\\ Khai Le-Duc$^{1}$, Weiyue Wang$^{2}$, Ralf Schl\"uter$^{1,2}$, Hermann Ney$^{1,2}$}

\address{$^1$Machine Learning and Human Language Technology, RWTH Aachen University, 52072 Aachen, Germany\\
$^2$AppTek GmbH, 52062 Aachen, Germany\\
   Email: \texttt{\{luescher,zeineldeen,zyang,raissi,vieting,schlueter,ney\}@cs.rwth-aachen.de}
}

\begin{document}

\graphicspath{{./gfx}}

\maketitle

\begin{abstract}
    Language barriers present a great challenge in our increasingly connected and global world.
Especially within the medical domain, e.g. hospital or emergency room, communication difficulties, and delays may lead to malpractice and non-optimal patient care.
In the HYKIST project, we consider patient-physician communication, more specifically between a German-speaking physician and an Arabic-, Vietnamese-, or Ukrainian-speaking patient.
Currently, a doctor can call the Triaphon service to get assistance from an interpreter in order to help facilitate communication.
The HYKIST goal is to support the usually non-professional bilingual interpreter with an automatic speech translation system to improve patient care and help overcome language barriers.
In this work, we present our ASR system development efforts for this conversational telephone speech translation task in the medical domain for two language pairs, data collection, various acoustic model architectures, and dialect-induced difficulties.

\end{abstract}

\section{Introduction}
\label{sec:intro}

In our globally connected world, it is becoming increasingly common to migrate to foreign countries, whether for work, refugee movements, or other reasons.
Consequently, language barriers between locals and foreigners pose a frequent everyday problem.
This can be particularly challenging for patient-physician communication during anamnesis, especially in the emergency room, leading to potential degradation of patient care.
The German Triaphon service\footnote{\href{https://triaphon.org}{https://triaphon.org}} is designed to help with communication in these situations, where no common language is shared between patient and physician.

The service provides a usually non-professional bilingual interpreter to help with patient-physician communication.
Nevertheless, communication regarding medical terms remains critical since these might not be known by the (medically not trained) interpreter, and the impact of erroneous or delayed translation might be particularly severe here.
In the HYKIST\footnote{\href{https://www.bundesgesundheitsministerium.de/ministerium/ressortforschung-1/handlungsfelder/forschungsschwerpunkte/digitale-innovation/modul-4-smarte-kommunikation/hykist.html}{HYKIST Project Webpage}} project, the goal is to develop a speech translation system to support the interpreter with automatic speech translation.
In this context, several challenges arise, like varying acoustic and recording conditions due to different devices and changing environments as well as bilingual input through a single recording channel, since the phone in the emergency room is shared by the physician and patient.
Additionally, we deal with telephone speech, as the interpreter is called via phone.

In this paper, we present our efforts on building \gls{ASR} systems for transcribing the telephone speech encountered in this project.
We collect data of (simulated) patient-physician conversations assisted by a Triaphon interpreter and create manual annotations.
Results are presented on all four languages considered in the HYKIST project, i.e., Arabic, German, Vietnamese, and Ukrainian.

\section{Related Work}
\begin{table*}[!t]
\centering
\caption{Data statistics for labeled telephone conversational speech data. Diversity level specifies the variability in the acoustic conditions of the speech data.}
\label{table:acousticdata}
\begin{tabular}{|l|c|c|r|r|c|c|c|}
\hline
Language                    & Dataset                 & Usage   & \multicolumn{1}{c|}{\# Spks} & \multicolumn{1}{c|}{Hours}            & Medical Domain       & Target-domain match   & Diversity level       \\ \hline\hline
\multirow{4}{*}{Arabic}     & In-house                & train   & 3379                         & 786                                   & no                   & Medium                & Medium                \\ \cline{2-8}
                            & \multirow{3}{*}{HYKIST} & adapt   & 3                            & 2                                     & \multirow{3}{*}{yes} & \multirow{3}{*}{High} & \multirow{3}{*}{Low}  \\ \cline{3-5}
                            &                         & dev     & 4                            & 3                                     &                      &                       &                       \\ \cline{3-5}
                            &                         & test    & 6                            & 3                                     &                      &                       &                       \\ \hline\hline
\multirow{4}{*}{German}     & In-house                & train   & 1723                         & 177                                   & no                   & Medium                & Medium                \\ \cline{2-8}
                            & \multirow{3}{*}{HYKIST} & adapt   & 5                            & 5                                     & \multirow{3}{*}{yes} & \multirow{3}{*}{High} & \multirow{3}{*}{Low}  \\ \cline{3-5}
                            &                         & dev     & 6                            & 3                                     &                      &                       &                       \\ \cline{3-5}
                            &                         & test    & 10                           & 3                                     &                      &                       &                       \\ \hline\hline
\multirow{4}{*}{Vietnamese} & In-house                & train   & 2240                         & 219                                   & no                   & Medium                & Medium                \\ \cline{2-8}
                            & \multirow{3}{*}{HYKIST} & adapt   & 1                            & 1                                     & \multirow{3}{*}{yes} & \multirow{3}{*}{High} & \multirow{3}{*}{Low}  \\ \cline{3-5}
                            &                         & dev     & 3                            & 3                                     &                      &                       &                       \\ \cline{3-5}
                            &                         & test    & 2                            & 2                                     &                      &                       &                       \\ \hline\hline
\multirow{3}{*}{Ukrainian}  & In-house                & train   & 1593                         & 407                                   & no                   & Medium                & Medium                \\ \cline{2-8}
                            & \multirow{2}{*}{HYKIST} & dev     & 37                           & 2                                     & \multirow{2}{*}{yes} & \multirow{2}{*}{High} & \multirow{2}{*}{High} \\ \cline{3-5}
                            &                         & test    & 39                           & 2                                     &                      &                       &                       \\ \hline
\end{tabular}
\end{table*}

\label{sec:related}
There has been researched work specifically addressing tasks within the medical domain.
A common issue for medical domain \gls{ASR} is challenging acoustic conditions and a privacy-related lack of domain-specific data \cite{edwards2017medicalspeech, chiu2018medconv, kar2021operation}.
The domain-specific medical terminology presents another challenge.
In \cite{sakti2014towards}, a multilingual system for the medical domain is presented.
Another approach to handle the medical domain is correcting \gls{ASR} errors on the output level \cite{mani2020towardsmedical}.
Earlier works presented \gls{ASR} systems for the four target languages, e.g. a system for Arabic in \cite{shaik2015arabic}.
German \gls{ASR} systems were developed within the Quaero project \cite{sundermeyer2011quaero} and similarly, Vietnamese was part of the Babel project \cite{golik2017babel_kws}. %

\section{Data}
\label{sec:data}

\subsection{HYKIST data}

Within the HYKIST project, our partner Triaphon recorded the Arabic-German and Vietnamese-German trialogues,
while we recorded the Ukrainian-German trialogues,
i.e. conversations between three persons: patient and physician on the emergency room end of the line using the same phone device,
and an interpreter on the other end of the line.
The patient speaks the non-German language -- Arabic or Vietnamese -- while the doctor speaks German.
The interpreter speaks both languages and helps the patient and doctor communicate.
Finding suitable speakers for each role was a challenging task since each role has a limiting factor on availability.
The physician speaker role requires medical training, limiting the speaker to physicians, doctors, and nurses.
While the patient role has the lowest bar on skill set, contacting and finding people with no or little German language skills in Germany was challenging.
The interpreter role also posed challenges since the role requires bilingualism.

For the Ukrainian-German trialogues, we recruited doctors and nurses from the hospitals in Aachen and established connections with Ukrainian refugees.
We used the Triaphon telephone service system to conduct the recordings.
The speakers were instructed to speak as normally as possible, in order to facilitate a natural conversation with interruptions, disfluencies, and hesitations.
Further, the physician was instructed to embed medical terms into the conversation.

The audio recordings of the simulated conversations were manually transcribed within the project.
The transcribers were provided with an initial transcription guideline, which was extended and modified with the help of the transcribers based on the encountered speech.

The data has been split into three subsets for each language: development and test data, as well as adaptation data, ensuring no speaker overlap between the sets.
In \tab\ref{table:acousticdata}, the data statistics are shown for the adapt, dev, and test sets for each individual language.
As shown the adapt, dev, and test set data from the HYKIST recordings are very limited.
For the dev and test sets the number of available hours is on the lower end.
To further complicate the HYKIST data, the number of speakers is limited, resulting in a low diversity level of the testing data.
This is due to the lack of speakers who can participate since each speaker's role requires an uncommon skill set.
Therefore we sourced additional training data from our industry partner AppTek and other sources to tackle the impact of the data issues.

\subsection{In-house data}

Our industry partner AppTek provided us with a decent amount of general annotated \SI{8}{\kilo\hertz} sampling rate conversational telephone speech data.
We are restricted to \SI{8}{\kilo\hertz} sampling rate due to the requirement of access over plain telephone services.
The audio data consists of telephone conversations of customers with varying call centers.
In \tab\ref{table:acousticdata}, the data statistics for the in-house training sets for each of the four languages are shown.
We can see that we have a discrepancy in the amount of in-house training data between the languages.
Especially, the amount of available German and Vietnamese data is very low.
For Ukrainian, the available data set is medium-sized.
Arabic has the largest amount of data.
Additionally, for the Arabic and Vietnamese data, we have speakers with accents and/or dialects.
For the Arabic data, we received four distinct datasets with different dialects: Syrian, Lebanese, Gulf, and Egyptian.
In the Vietnamese data, the speakers with accents are combined into one single dataset.

\subsection{Monolingual text data}

Our project partner AppTek provided monolingual text data for each of the four languages.
The data includes text from various sources.
In \tab\ref{table:lms}, the number of running words for each language is shown.
The available data amount varies between the languages, favoring German and Ukrainian while Arabic is at the lower end.

\subsection{Domain}

As shown in \tab\ref{table:acousticdata}, the data covers various domains.
The target domain of the HYKIST project is medical conversational telephone speech.
This specific domain is not covered by the in-house training data, which covers conversational telephone speech.
By listening to a number of randomly chosen audios and comparing them to our target domain,
we subjectively determined target-domain match and diversity level and ranked them as low, medium, and high.
The target-domain match specifies if the speech data matches our HYKIST scenario.
The diversity level indicates if the speech data covers varying acoustic conditions, like speakers, accents, noises, etc.

\section{Methods}
\subsection{Lexicon}

The initial pronunciation lexicon for the Vietnamese language was taken from the Babel project\footnote{\href{https://www.iarpa.gov/research-programs/babel}{https://www.iarpa.gov/research-programs/babel}}.
The initial German lexicon was sourced from the Quaero project\footnote{\href{http://www.quaero.org/}{http://www.quaero.org/}}.
For the Arabic system, the initial lexicon was provided by AppTek.
With the toolkit Sequitur Grapheme-To-Phoneme\footnote{\href{https://github.com/sequitur-g2p/sequitur-g2p}{https://github.com/sequitur-g2p/sequitur-g2p}} \cite{bisani2008g2p}, the initial lexica are extended, creating the training lexicon.

For decoding the HYKIST data, we further augment the lexicon with medical terms.
We acquired different sources for the medical terms.
Some medical terms were provided by our project partner Triaphon.
We scraped the web for more medical terms from medical websites and literature.
The generation of phoneme sequences for these medical terms proved challenging since the pronunciation is sometimes not clear.
For example, how are Latin or English (medical) terms pronounced in each language?
Solving these issues requires an understanding of the language itself and is sometimes not clearly answerable.
The final recognition lexica have a size of 370k words for Arabic, 202k for German, and 11k for Vietnamese, see~\tab\ref{table:lms}.

\subsection{Acoustic model}

To train the acoustic model, we follow the training pipeline described in \cite{luescher2019librispeech} to establish a \gls{GMM}-\gls{HMM} baseline for each of the languages.
This model is used to obtain an alignment of the speech data with the labels.
The \gls{NN} used for the hybrid model is trained on these alignments in a supervised way using the \gls{fCE} loss, following the setup from \cite{luescher2019librispeech,zeineldeen2022robustconformer}.
The experimental details for these models are described in \sect\ref{sec:experiments}.

\subsection{Language model}

The \glspl{LM} employed all use full-words and are count-based 4-grams.
We follow the same training recipe for all languages using the SRILM toolkit \cite{stolcke2002srilm} to build our \glspl{LM} using modified Kneser-Ney smoothing.
The initial step is to build a \gls{LM} for each individual monolingual text corpus.
Afterwards we evaluate the \glspl{LM} on the dev set and compute the weight for each \gls{LM} separately.
The weights determine the contribution of each individual \gls{LM} to the combined \gls{LM}.
The final step is to prune the \gls{LM} in order to reduce its size, resulting in one \gls{LM} per language.

\section{Experimental setups}
\label{sec:experiments}
In this section, we describe our experimental setups.
We use RETURNN\footnote{\href{https://github.com/rwth-i6/returnn}{https://github.com/rwth-i6/returnn}} \cite{doetsch2016returnn} for supervised training.
Decoding is performed with RASR\footnote{\href{https://github.com/rwth-i6/rasr}{https://github.com/rwth-i6/rasr}} \cite{rybach2011rasr}
We plan to publish training and decoding configurations online\footnote{\href{https://github.com/rwth-i6/returnn-experiments}{https://github.com/rwth-i6/returnn-experiments}}
\footnote{\href{https://github.com/rwth-i6/i6-experiments}{https://github.com/rwth-i6/i6-experiments}}.

The training recipe for the baseline systems for each language is similar and only varies in detail.
For all models, we generate alignments via a Gaussian mixture hidden Markov model process to be used as labels for neural network training.
All models are trained from scratch in a supervised manner with \gls{fCE}.
The acoustic model labels are generalized triphone states obtained by a \glsentrylong{CART} with 4501 labels.
As input to the acoustic model, we use 40-dimensional Gammatone features \cite{schlueter:icassp07}.
The learning rate schedule uses a warm-up phase followed by a hold phase and an exponential decay phase.
SpecAugment \cite{park2019specaugment} is applied during all model training.
We employ and compare different neural acoustic model architectures: \gls{BLSTM} \gls{RNN} \cite{hochreiter1997long}, \transformer \cite{vaswani2017attention} and \conformer \cite{google2020conformer} encoders.
For the \gls{BLSTM} model, we closely follow the training recipe in \cite{luescher2019librispeech}.
The \gls{BLSTM} employs five layers with 512 units per direction.
Our training recipe for the \transformer and \conformer is taken from \cite{zeineldeen2022conformer, zeineldeen2022robustconformer}.
The \transformer and \conformer contain 12 blocks each.
This results in different \gls{AM} sizes: \gls{BLSTM} 25M parameters, \transformer 90M parameters, and \conformer 53M parameters.
We train all models until convergence.
For the Arabic \gls{AM} training, we notice that removing the Egyptian dialect from training data helps which is probably due to the considerable difference between the Egyptian dialect and other dialects.
In addition, we observe that the quality of the alignments generated using \gls{GMM} system is bad.
Thus, we generate better alignments using the \gls{BLSTM} model
trained with the Gaussian mixture HMM alignment.
We apply another training iteration by training the \gls{BLSTM} model with
the better alignments and then generate alignments again.
These alignments are used to train the final best \conformer acoustic model.

\section{Experimental results}
In \tab\ref{table:lms}, we can see that the \glspl{LM}' performance varies drastically.
\Gls{OOV} rate gives the percentage of words with in the evaluation corpus which are not within the lexicon and therefor can not be recognized.
The Arabic and German vocabulary results in \gls{OOV} rates between $1.0\%$ and $1.6\%$ for the evaluation data sets.
The Vietnamese vocabulary covers the dev and test set very well with a \gls{OOV} rate of $0.1\%$ and $0.2\%$ respectively.
In contrast, the Ukrainian vocabulary does not cover the dev and test set as well, resulting in a \gls{OOV} of $5.9\%$ and $7.7\%$ respectively.
The evaluation metric \gls{PPL} informs how well a \gls{LM} fits to a given evaluation dataset.
The German and Vietnamese \glspl{LM} perform much better compared to the Arabic and Ukrainian \glspl{LM}.
The issue with Arabic LM training is that the medical-domain text data is written in Modern Standard Arabic, but the Arabic dialects have different spelling and words which lead to perplexity degradation.
The high Ukrainian \gls{OOV} rate lead to a \gls{PPL} degradation.
Additionally, some Ukrainians speak a mix of Ukrainian and Russian leading to further  \gls{PPL} degradation.

\begin{table}[!ht]
\setlength{\tabcolsep}{0.4em}
\centering
\caption{\Glspl{LM} for all four languages on the corresponding HYKIST dev and test sets. All \glspl{LM} are 4-grams.}
\label{table:lms}
\scalebox{0.96}{
\begin{tabular}{|l|r|r|c|r|c|r|}
\hline
\multirow{2}{*}{Language} & \multicolumn{1}{c|}{\# words}  & \multicolumn{1}{c|}{vocab} & \multicolumn{2}{c|}{dev}                & \multicolumn{2}{c|}{test}               \\ \cline{4-7}
                          & \multicolumn{1}{c|}{ in train} & \multicolumn{1}{c|}{size}  &        OOV         &        PPL         &        OOV         &        PPL         \\ \hline\hline
Arabic                    & 112M                           & 370k                       &        1.6\%       &        972         &        1.0\%       &        972         \\ \hline
German                    & 2400M                          & 202k                       &        1.4\%       &         72         &        1.6\%       &         78         \\ \hline
Vietnamese                & 500M                           & 11k                        &        0.1\%       &         67         &        0.2\%       &         69         \\ \hline
Ukrainian                 & 1100M                          & 313k                       &        5.9\%       &        492         &        7.7\%       &        669         \\ \hline
\end{tabular}
}
\end{table}

Each language-specific baseline is trained with the according in-house monolingual \SI{8}{\kilo\hertz} sampled telephone data.
\tab\ref{table:baselines} shows the performance of the \gls{ASR} systems.
The systems for each language show varying performance due to the inherent difficulty of the language and the data.

\begin{table}[!ht]
\centering
\caption{\Glspl{WER} [\%] for baselines on HYKIST data. Trained in a purely supervised manner on the monolingual in-house data.}
\label{table:baselines}
\scalebox{0.93}{
\begin{tabular}{|l|l|c|c|cc}
\hline
\multirow{2}{*}{Language}   & \multicolumn{1}{c|}{\multirow{2}{*}{AM}} & \multicolumn{2}{c|}{WER {[}\%{]}} & \multicolumn{2}{c|}{CER[\%]} \\ \cline{3-6}
                            & \multicolumn{1}{c|}{}                    &       dev       &       test      & \multicolumn{1}{c|}{dev}  & \multicolumn{1}{c|}{test} \\ \hline\hline
\multirow{3}{*}{Arabic}     & Gauss. mix.                              &       73.2      &       77.0      & \multicolumn{1}{c|}{38.6} & \multicolumn{1}{c|}{40.5} \\ \cline{2-6}
                            & BLSTM                                    &       43.4      &       42.9      & \multicolumn{1}{c|}{22.6} & \multicolumn{1}{c|}{25.2} \\ \cline{2-6}
                            & \conformer                               &       36.8      &       40.4      & \multicolumn{1}{c|}{17.2} & \multicolumn{1}{c|}{21.3} \\ \hline
\multirow{3}{*}{German}     & Gauss. mix.                              &       39.1      &       38.7      &  &  \\ \cline{2-4}
                            & BLSTM                                    &       27.1      &       24.0      &  &  \\ \cline{2-4}
                            & \transformer                             &       26.0      &       22.3      &  &  \\ \cline{1-4}
\multirow{3}{*}{Vietnamese} & Gauss. mix.                              &       62.2      &       59.7      &  &  \\ \cline{2-4}
                            & BLSTM                                    &       32.9      &       38.4      &  &  \\ \cline{2-4}
                            & \transformer                             &       31.0      &       35.1      &  &  \\ \cline{1-4}
\multirow{2}{*}{Ukrainian}  & BLSTM                                    &       29.1      &       41.3      &  &  \\ \cline{2-4}
                            & \conformer                               &       26.1      &       39.5      &  &  \\ \cline{1-4}

\end{tabular}
}
\end{table}

Among the four languages, the German system shows the best performance, followed by Vietnamese, Ukrainian and lastly Arabic.
There are several reasons which contribute to this observation.
The German and Vietnamese transcriptions are of higher quality compared to the Arabic transcriptions.
Furthermore, the Vietnamese have accented speech, and the Arabic data includes several dialects which add complexity.

Otherwise, the performance trends of the \gls{ASR} systems for the four languages are similar.
Moving from a Gaussian mixture \gls{HMM} framework to hybrid \gls{HMM} framework with a recurrent \gls{NN} (\gls{BLSTM}) leads to a relative \gls{WERR} of 35-44\% on HYKIST test.
Changing the neural acoustic encoder topology to incorporate self-attention mechanism (\transformer and \conformer) reduces the \gls{WER} by a 6-8\% relative on HYKIST test.

Arabic is a morphologically rich language.
Especially for Arabic dialect data, many words can be written in
different ways which can lead to high substitution word errors.
Thus, \tab\ref{table:baselines} shows \glspl{CER} for Arabic on HYKIST dev and test sets.
It achieves $17.2\%$ and $21.3\%$ \gls{CER} on dev and test sets, respectively.

\begin{table}[ht]
    \centering
    \caption{WERs [\%] and LM PPL for Arabic \conformer AM on HYKIST data after
    mapping top frequence dialect words to MSA words.}
    \label{tab:ar_hykist_dialect_map}
\begin{tabular}{|c|cc|cc|}
\hline
\multirow{2}{*}{AM} & \multicolumn{2}{c|}{LM PPL}     & \multicolumn{2}{c|}{WER {[}\%{]}} \\ \cline{2-5}
                    & \multicolumn{1}{c|}{dev} & test & \multicolumn{1}{c|}{dev}   & test \\ \hline
\conformer           & \multicolumn{1}{c|}{972} & 972  & \multicolumn{1}{c|}{36.8}  & 40.4 \\ \hline
\hspace{3mm}+ Mapping           & \multicolumn{1}{c|}{\textbf{892}} & \textbf{896}  & \multicolumn{1}{c|}{\textbf{35.8}}  & \textbf{38.9} \\ \hline
\end{tabular}
\end{table}

One issue with dialect Arabic is that it is difficult to find a large amount
of text data to train a good LM, especially in the medical domain unlike for
the case of \gls{MSA}.
Thus, depending on the application, it is possible to do some kind of
mapping between dialect Arabic words and \gls{MSA} words to reduce the confusion
made by frequent dialect words.
However, mixing dialect and \gls{MSA} words in the output sentence is not
consistent anymore but could work for other tasks such as speech translation
where we mainly care about the translated output at the end.
In general, this mapping approach would work only if all components
of the final system are trained in a consistent way by using this mapping.
Thus, we conduct an experiment where we selected the top 200 frequent dialect
words and mapped them to \gls{MSA} words manually.
We train a new LM using this mapping and the PPLs are shown
in \tab\ref{tab:ar_hykist_dialect_map}.
Note that we interpolate with LMs trained on \gls{MSA} text data.
The WERs and CERs after applying this mapping are reported in
Table \ref{tab:ar_hykist_dialect_map}.
With that, we can observe that we can achieve $3\%$ and $4\%$ relative
improvement in terms of WER on dev and test sets respectively.

Finally, we can highlight the inherent difficulty of the task at hand.
Several factors contribute to this, including challenging acoustic conditions and telephony bandwidth recordings, background noise.
In addition, the medical domain adds complexity because of the scarcity of in-domain data, medical terms, usage of medical face masks, and distressed or emotional speakers.
The speakers' dialects and accents also contribute as well as the inherent difficulties of the Arabic languages.
Our future work will focus, among others, on providing and exploiting custom-recorded in-domain data in training, multilingual supervised training as well as a multilingual decoder.
Moreover, we use unsupervised methods and large pre-trained models and present the results in \cite{rwth2022hykist_pretraining}.

\section{Conclusion}
In this work, we present our efforts on building \gls{ASR} systems for conversational telephone speech in the medical domain for two language pairs.
The patient-physician conversations in the HYKIST project, where patient and physician do not share a common language, are supported by non-professional bilingual interpreters, in order to facilitate medical anamnesis.
The recorded project-specific speech data comprised of simulated conversations are described along with additional data sources which can be exploited for developing \gls{ASR} systems.
Challenges within the project are the acoustic conditions in a hospital, for example, the emergency room, accents, and dialects, as well as the medical terms used during patient-physician conversations.
We present and compare supervised baselines using different acoustic encoder topologies within the hybrid \gls{HMM} framework.
Additionally, we present a dialect mapping method to handle the challenges presented by the Arabic dialects.

\section{Acknowledgements}
\label{sec:acknowledgement}

This work was partially supported by the project HYKIST funded by the German Federal Ministry of Health on the basis of a decision of the German Federal Parliament (Bundestag) under funding ID ZMVI1-2520DAT04A.\par
We thank Wilfried Michel, Eugen Beck, and Evgeny Matusov for their support in constructing the Ukrainian lexicon.
We thank our transcribers for their work in providing annotated audio data: Ahmed Ali, Ali Ismail, Alan Alrechah, Majd Hamza, Lavanya Colombus, Anna Schröder, Anna Vogelsang, Phuong Thanh Ngo, Minh Trang Nguyen, Thuy An Nguyen.
For recording and/or transcribing the Ukrainian trialogues, we thank: Irina Nasibian, Igor Guci, Yulia Kostadinova, Olha Fryndak, Oleg Guzik, and Hendrick Alkemade.

\small
\bibliographystyle{ieeetr}
\bibliography{mybib}

\begin{thebibliography}{10}

\bibitem{edwards2017medicalspeech}
E.~Edwards, W.~Salloum, G.~P. Finley, J.~Fone, G.~Cardiff, M.~Miller, and
  D.~Suendermann-Oeft, ``Medical speech recognition: Reaching parity with
  humans,'' in {\em Speech and Computer} (A.~Karpov, R.~Potapova, and
  I.~Mporas, eds.), (Cham), pp.~512--524, Springer International Publishing,
  2017.

\bibitem{chiu2018medconv}
C.-C. Chiu, A.~Tripathi, K.~Chou, C.~Co, N.~Jaitly, D.~Jaunzeikare, A.~Kannan,
  P.~Nguyen, H.~Sak, A.~Sankar, J.~Tansuwan, N.~Wan, Y.~Wu, and X.~Zhang,
  ``{Speech Recognition for Medical Conversations},'' in {\em Proc. Interspeech
  2018}, pp.~2972--2976, 2018.

\bibitem{kar2021operation}
S.~Kar, P.~Mishra, J.~Lin, M.-J. Woo, N.~Deas, C.~Linduff, S.~Niu, Y.~Yang,
  J.~McClendon, D.~H. Smith, M.~C. Smith, R.~W. Gimbel, and K.-C. Wang,
  ``Systematic evaluation and enhancement of speech recognition in operational
  medical environments,'' in {\em 2021 International Joint Conference on Neural
  Networks (IJCNN)}, pp.~1--8, 2021.

\bibitem{sakti2014towards}
S.~Sakti, K.~Kubo, S.~Matsumiya, G.~Neubig, T.~Toda, S.~Nakamura, F.~Adachi,
  and R.~Isotani, ``Towards multilingual conversations in the medical domain:
  Development of multilingual medical data and a network-based {ASR} system,''
  in {\em Proc LREC}, (Reykjavik, Iceland), pp.~2639--2643, May 2014.

\bibitem{mani2020towardsmedical}
A.~Mani, S.~Palaskar, and S.~Konam, ``Towards understanding {ASR} error
  correction for medical conversations,'' in {\em Proc. NLPMC}, pp.~7--11,
  2020.

\bibitem{shaik2015arabic}
M.~A.~B. Shaik, Z.~T{\"u}ske, M.~A. Tahir, M.~Nu{\ss}baum-Thom,
  R.~Schl{\"u}ter, and H.~Ney, ``Improvements in {RWTH} {LVCSR} evaluation
  systems for {P}olish, {P}ortuguese, {E}nglish, {U}rdu, and {A}rabic,'' in
  {\em Proc. Interspeech}, 2015.

\bibitem{sundermeyer2011quaero}
M.~Sundermeyer, M.~Nu{\ss}baum-Thom, S.~Wiesler, C.~Plahl, A.~E.-D. Mousa,
  S.~Hahn, D.~Nolden, R.~Schl{\"u}ter, and H.~Ney, ``The {RWTH} 2010 {Q}uaero
  {ASR} evaluation system for {E}nglish, {F}rench, and {G}erman,'' in {\em
  Proc. ICASSP}, pp.~2212--2215, IEEE, 2011.

\bibitem{golik2017babel_kws}
P.~Golik, Z.~T\"uske, K.~Irie, E.~Beck, R.~Schl\"uter, and H.~Ney, ``The 2016
  {RWTH} keyword search system for low-resource languages,'' in {\em
  International Conference Speech and Computer}, (Hatfield, UK), pp.~719--730,
  Sept. 2017.

\bibitem{bisani2008g2p}
M.~Bisani and H.~Ney, ``Joint-sequence models for grapheme-to-phoneme
  conversion,'' {\em Speech Communication}, vol.~50, no.~5, pp.~434--451, 2008.

\bibitem{luescher2019librispeech}
C.~L\"uscher, E.~Beck, K.~Irie, M.~Kitza, W.~Michel, A.~Zeyer, R.~Schl\"uter,
  and H.~Ney, ``{RWTH} {ASR} systems for {LibriSpeech}: Hybrid vs
  {Attention},'' in {\em Proc. Interspeech}, (Graz, Austria), Sept. 2019.

\bibitem{zeineldeen2022robustconformer}
M.~Zeineldeen, J.~Xu, C.~L\"uscher, R.~Schl\"uter, and H.~Ney, ``Improving the
  training recipe for a robust conformer-based hybrid model,'' in {\em Proc.
  Interspeech}, (Incheon, Korea), Sept. 2022.

\bibitem{stolcke2002srilm}
A.~Stolcke, ``{SRILM} -- an extensible language modeling toolkit,'' in {\em
  Proc. ICSLP}, pp.~901--904, 2002.

\bibitem{doetsch2016returnn}
P.~Doetsch, A.~Zeyer, P.~Voigtlaender, I.~Kulikov, R.~Schl{\"u}ter, and H.~Ney,
  ``{RETURNN}: the {RWTH} extensible training framework for universal recurrent
  neural networks,'' in {\em Proc. ICASSP}, (New Orleans, {LA}, {USA}), 2017.

\bibitem{rybach2011rasr}
D.~Rybach, S.~Hahn, P.~Lehnen, D.~Nolden, M.~Sundermeyer, Z.~T{\"u}ske,
  S.~Wiesler, R.~Schl{\"u}ter, and H.~Ney, ``{RASR} - the {RWTH Aachen
  University} open source speech recognition toolkit,'' in {\em Proc. ASRU},
  (Waikoloa, HI, USA), Dec. 2011.

\bibitem{schlueter:icassp07}
R.~Schl{\"u}ter, I.~Bezrukov, H.~Wagner, and H.~Ney, ``Gammatone features and
  feature combination for large vocabulary speech recognition,'' in {\em Proc.
  ICASSP}, (Honolulu, HI, USA), pp.~649--652, Apr. 2007.

\bibitem{park2019specaugment}
D.~S. Park, W.~Chan, Y.~Zhang, C.-C. Chiu, B.~Zoph, E.~D. Cubuk, and Q.~V. Le,
  ``Specaugment: A simple data augmentation method for automatic speech
  recognition,'' {\em arXiv preprint arXiv:1904.08779}, 2019.

\bibitem{hochreiter1997long}
S.~Hochreiter and J.~Schmidhuber, ``Long short-term memory,'' {\em Neural
  computation}, vol.~9, no.~8, pp.~1735--1780, 1997.

\bibitem{vaswani2017attention}
A.~Vaswani, N.~Shazeer, N.~Parmar, J.~Uszkoreit, L.~Jones, A.~N. Gomez, L.~u.
  Kaiser, and I.~Polosukhin, ``Attention is all you need,'' in {\em Advances in
  Neural Information Processing Systems} (I.~Guyon, U.~V. Luxburg, S.~Bengio,
  H.~Wallach, R.~Fergus, S.~Vishwanathan, and R.~Garnett, eds.), vol.~30,
  Curran Associates, Inc., 2017.

\bibitem{google2020conformer}
A.~Gulati, J.~Qin, C.-C. Chiu, N.~Parmar, Y.~Zhang, J.~Yu, W.~Han, S.~Wang,
  Z.~Zhang, Y.~Wu, {\em et~al.}, ``Conformer: Convolution-augmented transformer
  for speech recognition,'' {\em arXiv preprint arXiv:2005.08100}, 2020.

\bibitem{zeineldeen2022conformer}
M.~Zeineldeen, J.~Xu, C.~L{\"u}scher, W.~Michel, A.~Gerstenberger,
  R.~Schl{\"u}ter, and H.~Ney, ``Conformer-based hybrid {ASR} system for
  switchboard dataset,'' in {\em Proc. ICASSP}, pp.~7437--7441, IEEE, 2022.

\bibitem{rwth2022hykist_pretraining}
P.~Vieting, C.~L\"uscher, J.~Dierkes, R.~Schl\"uter, and H.~Ney, ``Efficient
  use of large pre-trained models for low resource {ASR},'' in {\em {ICASSP}
  Workshop}, (Rhodes, Greece), IEEE, June 2023.

\end{thebibliography}

\end{document}